\documentclass[letterpaper]{article} 
\usepackage{aaai25}  
\usepackage{times}  
\usepackage{helvet}  
\usepackage{courier}  
\usepackage[hyphens]{url}  
\usepackage{graphicx} 
\usepackage{natbib}  
\usepackage{caption} 
\usepackage{algorithm}

\usepackage{newfloat}
\usepackage{listings}
\DeclareCaptionStyle{ruled}{labelfont=normalfont,labelsep=colon,strut=off} 
\floatstyle{ruled}
\newfloat{listing}{tb}{lst}{}
\floatname{listing}{Listing}
\title{Forecasting Fails: Unveiling Evasion Attacks in Weather Prediction Models}
\author {
    Huzaifa Arif\textsuperscript{\rm 1}\thanks{Huzaifa was supported by the Data Science Summer Internship (DSSI) program at LLNL.},
    Pin-Yu Chen\textsuperscript{\rm 2},
    Alex Gittens\textsuperscript{\rm 1},
    James Diffenderfer\textsuperscript{\rm 3},
    Bhavya Kailkhura\textsuperscript{\rm 3}
}
\affiliations {
    \textsuperscript{\rm 1}Rensselaer Polytechnic Institute, Troy, NY, United States\\
    \textsuperscript{\rm 2}IBM Research, Yorktown Heights, NY, United States\\
    \textsuperscript{\rm 3}Lawrence Livermore National Laboratory, Livermore, CA, United States\\
    arifh@rpi.edu, pin-yu.chen@ibm.com, gittea@rpi.edu, diffenderfer2@llnl.gov, kailkhura1@llnl.gov
}

\usepackage{bibentry}
\usepackage[dvipsnames]{xcolor}
\usepackage{comment}
\usepackage{amssymb} 
\usepackage{booktabs}
\usepackage{tabularx} 
\usepackage{algorithm}
\usepackage{algpseudocode}
\usepackage{caption}
\usepackage{subcaption}   

\usepackage[export]{adjustbox}

\usepackage{empheq}
\usepackage{fancybox}
\usepackage{colortbl}

\definecolor{mathcolor}{rgb}{0.8, 0.8, 0.8} 
\definecolor{textcolor}{rgb}{0.4, 0.4, 0.4} 

\newsavebox{\mysaveboxM} 
\newsavebox{\mysaveboxT} 

\definecolor{myblue}{rgb}{.8, .8, 1}

\definecolor{codegreen}{rgb}{0,0.6,0}
\definecolor{codegray}{rgb}{0.5,0.5,0.5}
\definecolor{codepurple}{rgb}{0.58,0,0.82}
\definecolor{backcolour}{rgb}{0.95,0.95,0.92}

\usepackage{amsthm}

\theoremstyle{definition}

\begin{document}

\maketitle

\begin{abstract}
With the increasing reliance on AI models for weather forecasting, it is imperative to evaluate their vulnerability to adversarial perturbations. This work introduces \textbf{Weather Adaptive Adversarial Perturbation Optimization (WAAPO)}, a novel framework for generating targeted adversarial perturbations that are both effective in manipulating forecasts and stealthy to avoid detection. \textbf{WAAPO} achieves this by incorporating constraints for channel sparsity, spatial localization, and smoothness, ensuring that perturbations remain physically realistic and imperceptible. Using the ERA5 dataset and FourCastNet~\cite{pathak2022fourcastnet}, we demonstrate \textbf{WAAPO}'s ability to generate adversarial trajectories that align closely with predefined targets, even under constrained conditions. Our experiments highlight critical vulnerabilities in AI-driven forecasting models, where small perturbations to initial conditions can result in significant deviations in predicted weather patterns. These findings underscore the need for robust safeguards to protect against adversarial exploitation in operational forecasting systems. 
The code for \textbf{WAAPO} is available at: https://github.com/Huzaifa-Arif/WAPPO
\end{abstract}

\section{Introduction}

Recent research has focused extensively on the development of artificial intelligence models for weather prediction tasks, leading to the creation of advanced AI-based prediction models such as FourcastNet \cite{pathak2022fourcastnet}, GraphCast \cite{lam2022graphcast}, ClimaX \cite{huang2023climax}, and PonguWeather \cite{bi2022pangu}. These models have demonstrated impressive accuracy and efficiency in weather forecasting, some of which can generate a 10-day forecast in just one minute \cite{rackow2024robustness}. Their accuracy often rivals that of traditional physics-based models, sparking significant interest from both private-sector companies and government agencies. The European Center for Medium-Range Weather Forecasts (ECMWF) and the National Oceanic and Atmospheric Administration (NOAA), which operates the Global Forecast System (GFS), have recognized the potential of these AI models. On 6 September 2023, ECMWF tweeted that three of its AI forecast models accurately predicted the slow westward movement of Hurricane Lee in the Atlantic Ocean, underscoring the practical application of these AI-driven forecasts \cite{hurricanelee}.

\begin{figure*}[t]
\centering
\begin{minipage}{0.95\linewidth}
    \centering
    \begin{subfigure}[t]{0.50\linewidth}
        \centering
        \includegraphics[width=\linewidth]{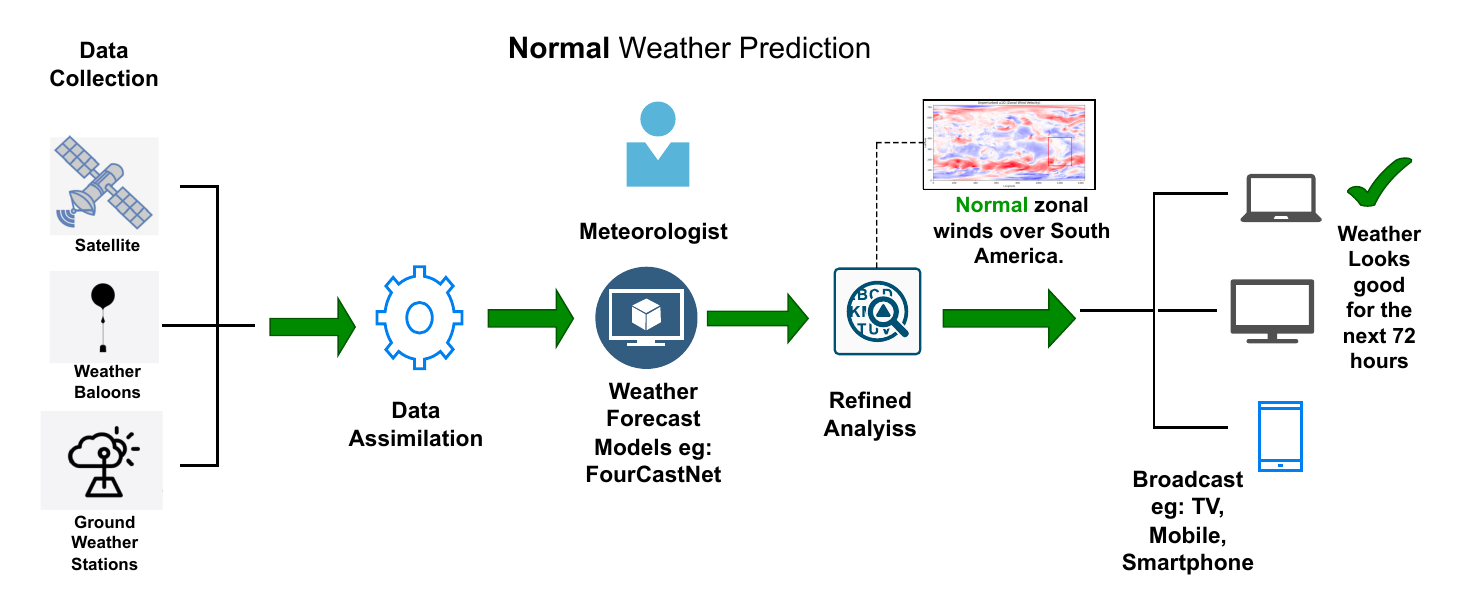}
        \caption{Unperturbed forecast process.}
        \label{fig:unperturb-process}
    \end{subfigure}
    \hfill
    \begin{subfigure}[t]{0.46\linewidth}
        \centering
        \includegraphics[width=\linewidth]{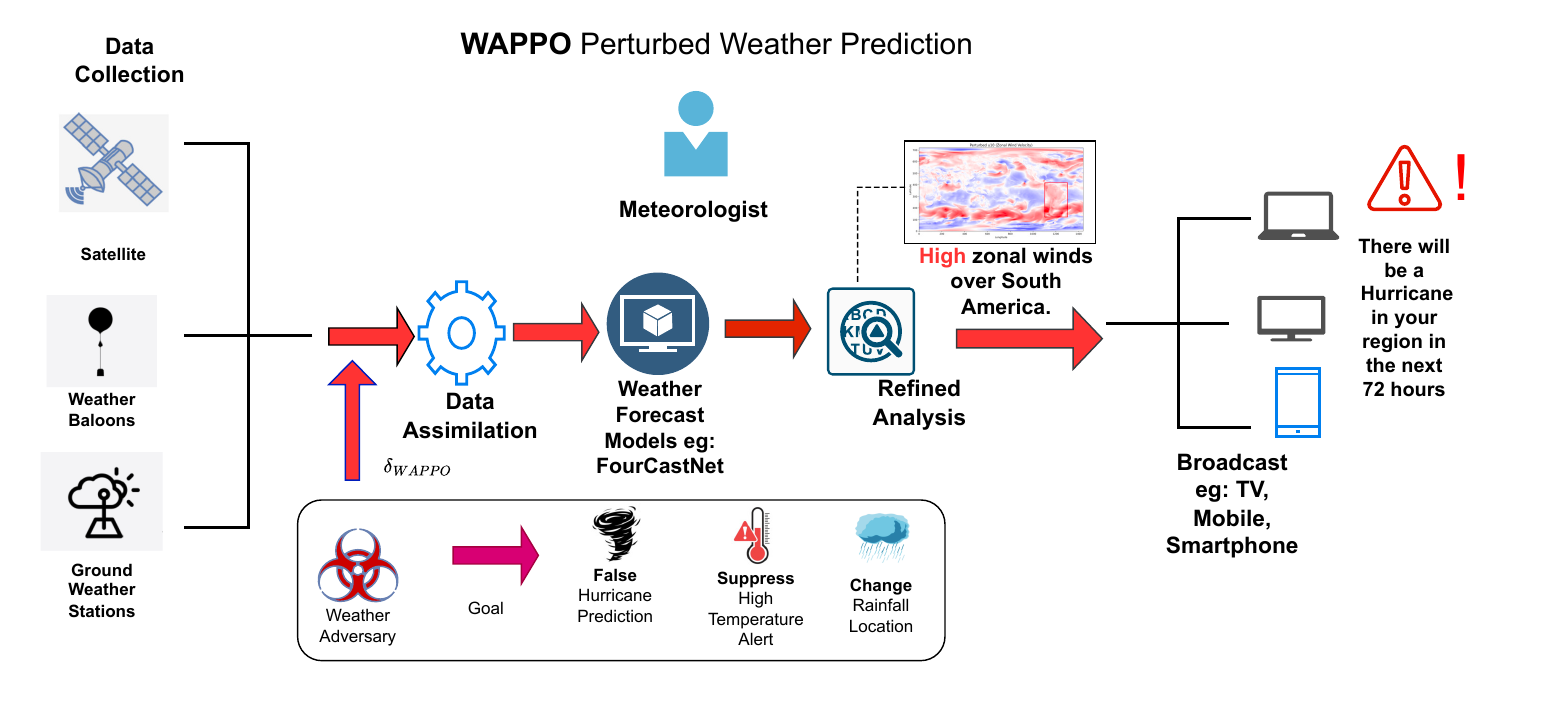}
        \caption{Perturbed forecast process.}
        \label{fig:perturb-process}
    \end{subfigure}
\end{minipage}
\caption{The weather forecasting process involves several key steps: \textbf{data collection, data assimilation, forecasting, analysis, and dissemination}. Data from sources like weather stations and satellites is processed through assimilation and forecasting models, refined through analysis, and shared with users via devices such as phones or TVs. Our study highlights vulnerabilities in this process, showing that adversaries can exploit the data collection phase to introduce perturbations and generate targeted false forecasts.}
\label{fig:teaser-diag}
\end{figure*}

Despite this success, limited research has been conducted to assess the vulnerability of these AI models to adversarial attacks. Traditional climate science institutes, such as ECMWF, employ rigorous protocols to ensure model reliability in prediction. Our study highlights a potential vulnerability by exploring whether these AI models could be susceptible to \textbf{ adverse manipulation of initial weather fields}. We introduce a novel problem in the context of weather forecasting, specifically targeting the inference phase of weather forecasting models, demonstrating that they are highly vulnerable to input perturbations (see Figure \ref{fig:teaser-diag}). 

Our study has significant implications, as creating false weather events or erasing or mistiming real ones could lead to serious consequences if these models are deployed without adversarial safeguards.


\subsection{Adversarial Scenario  in Weather Forecast Models}

Climate forecasting relies on sensor data that are vulnerable to cyberphysical attacks (see Figure~\ref{fig:teaser-diag}), enabling attackers to manipulate sensor data and inject inaccuracies into forecasting models. These models are often used by government-owned entities, such as NOAA. Even government-controlled sensors and models are not immune, as demonstrated by the following past incidents.

In September 2014, the National Oceanic and Atmospheric Administration (NOAA) experienced a significant cyberattack in which vulnerabilities in its IT security program were exploited. This attack compromised four NOAA websites and temporarily disrupted satellite imagery feeds, highlighting long-standing deficiencies in the NOAA cybersecurity infrastructure \cite{NOAA2014Cyberattack}. Similarly, in an unrelated application, in November 2024, China-linked hackers infiltrated multiple US telecommunications providers, extracting data on legal wiretaps and eavesdropping on government and political conversations. This breach, which affected roughly ten providers, including Verizon and AT\&T, marked a substantial counterintelligence failure \cite{Telecom2024Breach}.

These cases illustrate the tangible threat of cyberattacks on forecasting models, particularly as these models become more widely used and increasingly depend on data from independent sources, thereby expanding the attack surface. Attackers could exploit less secure third-party data providers to manipulate model inputs, potentially leading to inaccurate weather predictions. Such inaccuracies could result in economic losses, disrupted logistics, and public safety risks.

Therefore, vigilance and robust cybersecurity measures are critical to safeguarding forecasting systems against these potential threats.

\section{Related Works}
\begin{figure}[htbp]
    \centering
    \begin{tabular}{c c c}
        & \textbf{CaffeNet} & \textbf{VGG-F} \\[5pt]
        
        \rotatebox{90}{\textbf{Original}} &
        \includegraphics[width=0.4\linewidth]{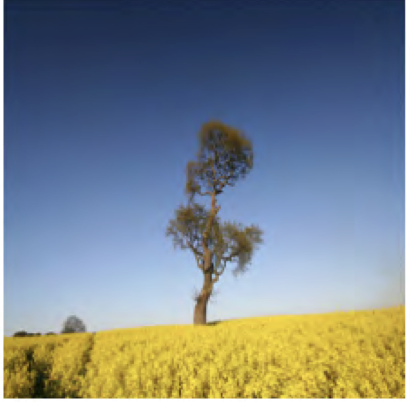} &
        \includegraphics[width=0.4\linewidth]{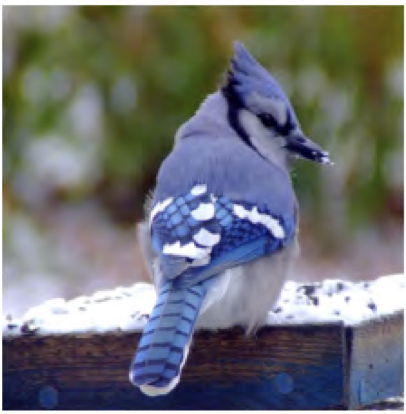} \\[5pt]
        & \textit{``rapeseed" 99.9\% confidence} & \textit{``jay" 99.9\% confidence} \\[10pt]
        
        \rotatebox{90}{\textbf{Perturbed}} &
        \includegraphics[width=0.4\linewidth]{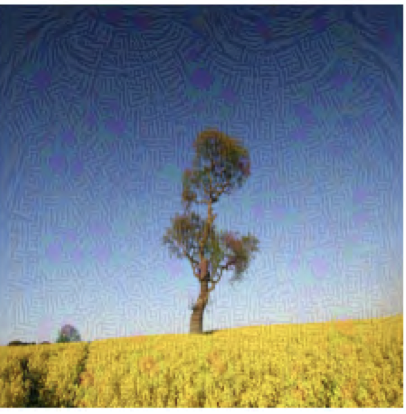} &
        \includegraphics[width=0.4\linewidth]{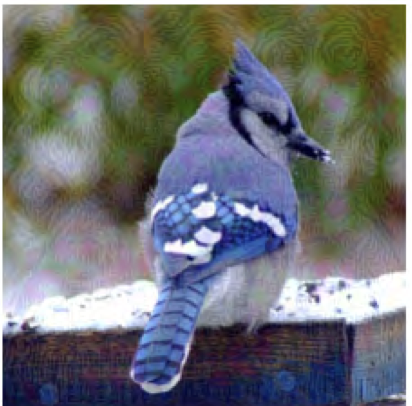} \\[5pt]
        & \textit{``cardigan" 89.7\% confidence} & \textit{``mask" 81.8\% confidence} \\
    \end{tabular}
    \caption{Comparison of Original and Perturbed Images for CaffeNet and VGG-F when the underlying target model is a classifier. An imperceptible perturbation leads to a different class with high confidence.}
    \label{fig:classifier examples}
\end{figure}

Typically, the literature focuses on adversarial attacks, where the target model is a classifier \cite{costa2024deep}. In such settings, the goal of the adversary is defined as:

\begin{equation}
\arg\min_{\delta \mathbf{X}} \|\delta \mathbf{X}\| \quad \text{s.t.} \quad f(\mathbf{X} + \delta \mathbf{X}) = \mathbf{Y}^*,
\end{equation}

\begin{equation}
\mathbf{X}^* = \mathbf{X} + \delta \mathbf{X},
\end{equation}

where \(f\) is the target classifier and \(\mathbf{Y}^*\) is the target class defined by the adversary. For a clean example \((\mathbf{X}, \mathbf{Y})\), the adversary seeks a perturbation \(\delta \mathbf{X}\) that results in the output \((\mathbf{X} + \delta \mathbf{X}, \mathbf{Y}^*)\). To maintain imperceptibility, the perturbation is constrained, typically using \(\ell_\infty\) or \(\ell_2\)-norms. This ensures that the perturbed input appears similar to the clean input, while causing the classifier to misclassify with high confidence. A visual depiction is provided in Figure~\ref{fig:classifier examples}, reproduced from \cite{akhtar2018threat}.

Numerous approaches to executing adversarial attacks in the classification setting have been proposed in the literature. Common methods include the box-constrained L-BFGS \cite{long2022survey}, Fast Gradient Sign Method (FGSM) \cite{goodfellow2014explaining}, and the Carlini and Wagner (C\&W) attack \cite{carlini2017towards}. Additional techniques are summarized in the surveys \cite{costa2024deep,akhtar2018threat}.

With the rise of generative models, there has been a growing interest in exploring adversarial attacks and defenses in this domain \cite{gen1,gen2}. These attacks are typically categorized as  \textit{untargeted} adversarial attacks \cite{untarget1}:
\begin{equation}
\left\{ \mathbf{X}_{\text{adv}} \ \middle|\ \|\mathbf{X}_{\text{adv}} - \mathbf{X}\|_p \leq \epsilon \ \text{and} \ f_\theta(\mathbf{X}) \neq f_\theta(\mathbf{X}_{\text{adv}}) \right\}
\label{eqn: untarget}
\end{equation}
or \textit{targeted} adversarial attacks \cite{target1}, where \(t_{\text{adv}}\) is a predefined adversarial target:

\begin{equation}
\left\{ \mathbf{X}_{\text{adv}} \ \middle|\ \|\mathbf{X}_{\text{adv}} - \mathbf{X}\|_p \leq \epsilon \ \text{and} \ f_\theta(\mathbf{X}_{\text{adv}}) = t_{\text{adv}} \right\}.
\label{eqn: target}
\end{equation}

In this setting, an autoregressive model is repeatedly queried with some initial input over a fixed time horizon. While adversarial attacks have been extensively studied in tasks like time-series forecasting, their application to weather forecasting remains largely unexplored in the literature. This work aims to address this gap by introducing a study of \textit{targeted} adversarial attacks specifically designed for weather forecasting models. Our analysis is conducted in the \textbf{white-box} setting, where the adversary has complete knowledge of the forecasting model and its parameters.

The objective of this novel adversarial attack strategy is to perturb the initial conditions of a weather forecasting model to cause significant deviations in the predicted trajectories for manipulating extreme weather events, such as hurricanes, toward some predefined target event. Importantly, these perturbations must satisfy constraints of physical realism and stealth, ensuring they remain relatively imperceptible, undetectable to unbounded perturbations and do not violate physical laws of weather fields.

Our experimental setup employs the state-of-the-art FourCastNet \cite{pathak2022fourcastnet,nguyen2024climatelearn} as the weather forecasting model, using the widely adopted ERA5 dataset for evaluation. The results demonstrate the effectiveness of this adversarial evasion attack in influencing weather forecasts by employing physically plausible and stealthy perturbations.

\begin{figure*}[t]
    \centering
    \includegraphics[width=0.9\linewidth]{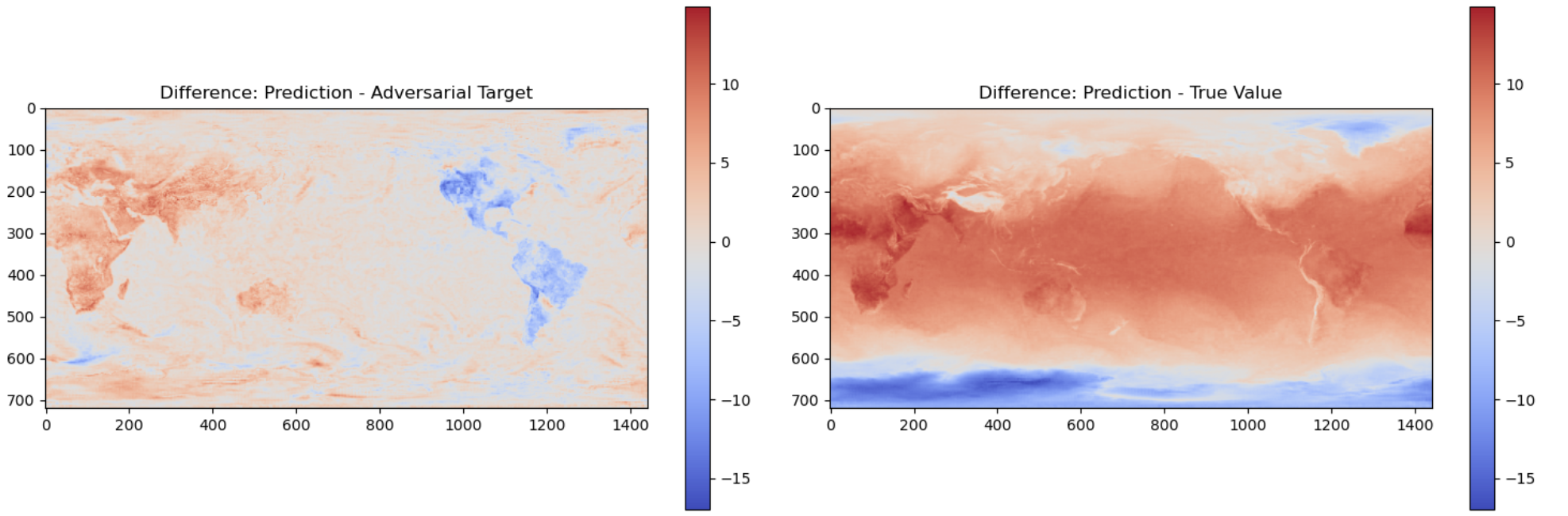}
    \caption{Pointwise temperature differences (in Kelvin) comparing the 24-hour perturbed forecast with both the 120-hour \textbf{adversarial target} ($t_{\text{adv}}$) and the 24-hour \textbf{ground truth (GT/True Value)}. Despite the adversarial target representing conditions 120 hours into the future, the unconstrained attack effectively manipulates the model’s prediction to align more closely with the adversarial target than the actual 24-hour ground truth. This highlights the attack's capability to fabricate a false global temperature event, significantly overriding the model’s original forecast.}
    \label{fig:pointwise-mse}
\end{figure*}

\begin{figure*}
    \centering
    \begin{subfigure}{0.48\textwidth}
        \centering
        \includegraphics[width=\linewidth]{figures/pointwise_mse.png}
        \caption{WAAPO applied to only the \textbf{temperature (t2m)} channel.}
        \label{fig:channel_only}
    \end{subfigure}
    \hfill
    \begin{subfigure}{0.48\textwidth}
        \centering
        \includegraphics[width=\linewidth]{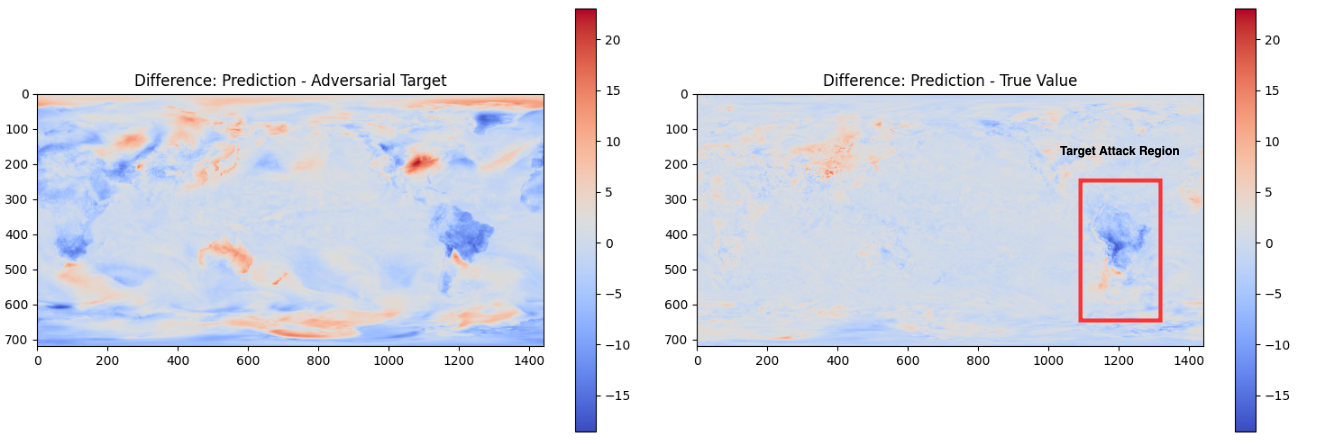}
        \caption{WAAPO  with a \textbf{spatial mask} over South America.}
        \label{fig:patch_only}
    \end{subfigure}
    \caption{These results illustrate the \textbf{Weather Adaptive Adversarial Perturbation Optimization (WAAPO)} framework from Algorithm~\ref{alg:adversarial_perturbation}. In \textbf{(a)}, WAAPO targets only the temperature ($t2m$) channel (Kelvin), showing that even single-channel perturbations can significantly alter predictions, closely mirroring the behavior seen in Figure~\ref{fig:pointwise-mse}. In \textbf{(b)}, a spatial mask $M$ is applied over South America, demonstrating that localized, channel-specific perturbations can substantially reshape forecasts within the targeted region. }
    \label{fig:combined_results}
\end{figure*}




\section{Problem Statement}


An adversary can introduce subtle, realistic modifications to the \textbf{initial} weather conditions, mirroring plausible real-world scenarios, to manipulate weather forecasts toward desired outcomes. The practical implications of such perturbations are detailed in  Appendix. For instance, if the adversary aims to create a \textbf{false temperature} event, they might adjust temperature profiles to trigger alerts for heat measures in a targeted region. By altering initial temperature fields, they could change the predicted temperature distribution across a region, affecting forecasts for heatwaves or cold spells. Each of these objectives relies on small, carefully controlled adjustments to particular atmospheric variables, allowing the adversary to steer forecasts toward their desired outcomes.


Table \ref{tab:variables} provides an overview of the notations used in this paper. The autoregressive AI forecast model (e.g., FourCastNet \cite{pathak2022fourcastnet}) is denoted by $\phi$, which generates a trajectory of forecasts $\mathbf{Z} \in \mathbb{R}^{T \times L \times M \times N}$ over a time horizon $T$. Here, $N$ represents the prognostic variables (e.g., Temperature, Surface Wind Speed, etc.; see Table \ref{tab:prognostic_variables} for more details on the prognostic variables used by FourCastNet). $L$ and $M$ denote the number of latitude and longitude points, respectively (e.g., $L = 721$, $M = 1440$ for FourCastNet \cite{pathak2022fourcastnet}).

We use the subscript $i$ to indicate the forecast at a specific time point. For instance, $\mathbf{Z}_3$ represents the model's third prediction for a given initial condition. The initial condition $\mathbf{Z}_0 \in \mathbb{R}^{L \times M \times N}$ is provided as input to $\phi(\cdot)$, which then generates the sequence of predictions $\mathbf{Z}_{1:T}$ which for simplicity is replaced by $\mathbf{Z}$ when discussing the whole forecast trajectory.


\begin{table}[htbp]
    \centering
    \caption{Definitions of Variables}
    \label{tab:variables}
    \small 
    \begin{tabularx}{\columnwidth}{@{} l X @{}}
        \toprule
        \textbf{Symbol} & \textbf{Description} \\ 
        \midrule
        $\phi(\cdot)$ & Weather forecast model (e.g., ForecastNet) \\
        $\mathbf{Z}$ & Set of \( T \) future forecast values, of shape $T \times L \times M \times N$ \\
        $\mathbf{Z}_0$ & Initial condition, of shape $L \times M \times N$ \\
        $\delta$ & Perturbation applied to the initial conditions \\
        $t_{\text{adv}}$ & Desired adversarial event, of shape $L \times M \times N$\\
        $L$ & Longitude \\
        $M$ & Latitude \\
        $N$ & Number of prognostic variables (e.g., temperature, wind speed) \\
        \bottomrule
    \end{tabularx}
\end{table}

In this study, we address the challenge of designing a perturbation \( \delta \in \mathbb{R}^{L \times M \times N} \) applied to the initial conditions \( \mathbf{Z}_0 \) to manipulate a weather forecast model \( \phi(\cdot) \) (e.g., ForecastNet). Specifically, our objective is to minimize the squared \( \ell_2 \)-norm difference between the model's forecast at time \( T \), denoted as $\mathbf{Z}_T = \phi_{T}(\mathbf{Z}_0 + \delta)$, and a desired adversarial future event \( t_{\text{adv}} \). Here the subscript \(T\) of the model \(\phi\) denotes the prediction of the model at the $T^{\text{th}}$ timestep. Given a predetermined trajectory length \( T \), the goal is to design the perturbation \( \delta \) such that, when added to the initial condition, it causes the forecast to deviate significantly within the time frame \( T \), making it resemble \( t_{\text{adv}} \). This approach constitutes a \textbf{targeted} evasion attack for weather forecast task. 

\section{Are Weather Forecasts Secure to Unconstrained Attacks?}

\begin{figure*}[t]
    \centering
    \begin{subfigure}[b]{0.9\linewidth}
        \centering
        \includegraphics[width=\linewidth]{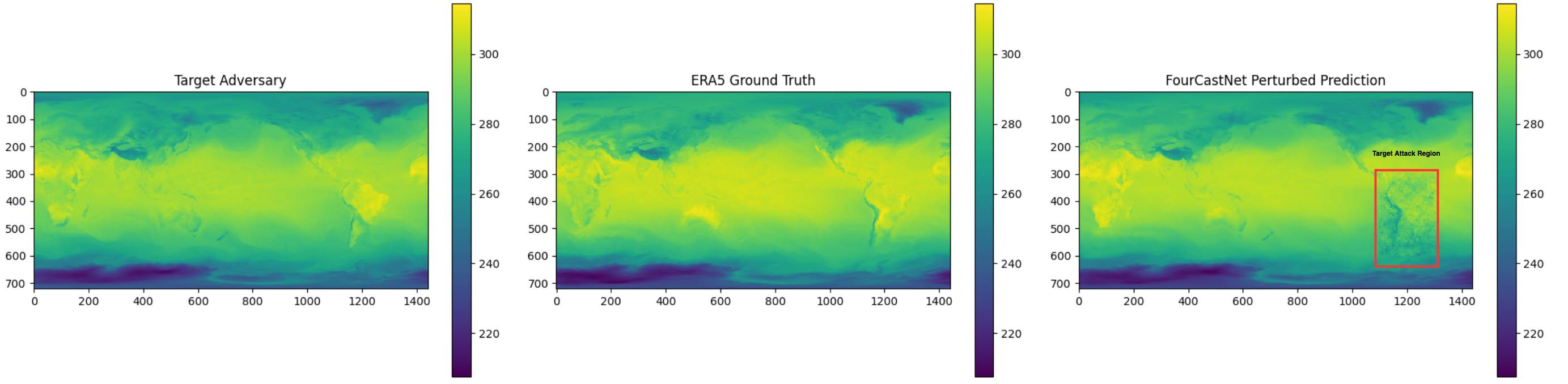}
        \label{fig:non-smooth}
    \end{subfigure}
    \hfill
    \begin{subfigure}[b]{0.9\linewidth}
        \centering
        \includegraphics[width=\linewidth]{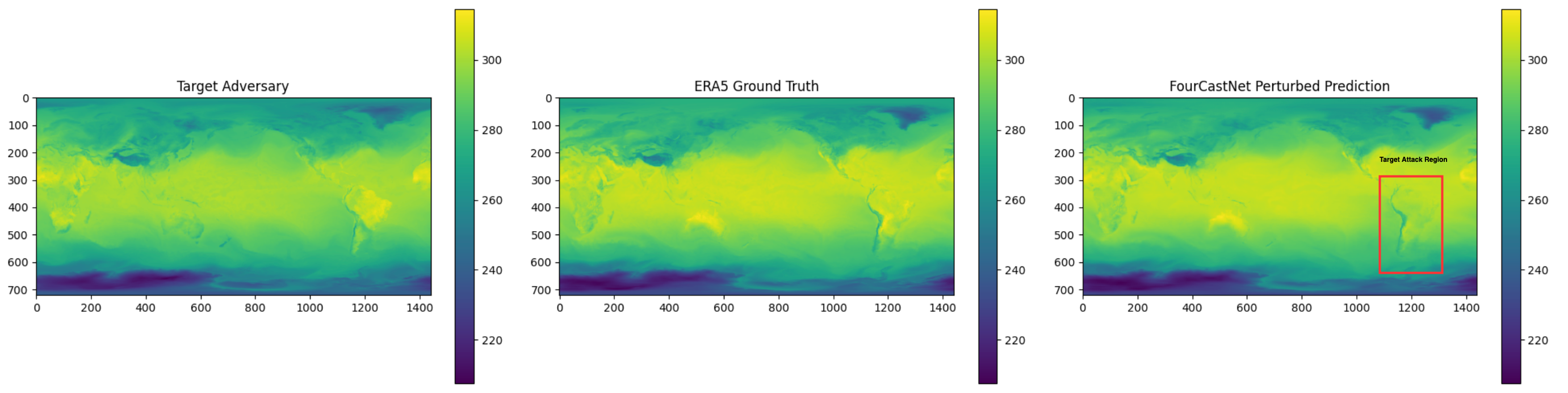}
        \label{fig:smooth}
    \end{subfigure}
    \caption{This figure demonstrates why a \textit{smoothness constraint} is essential for creating imperceptible perturbations using \textbf{WAPPO}. The top row, without smoothness constraints, displays a clearly visible patch in the targeted area, whereas applying a smoothness constraint in the bottom row yields more diffused, subtle perturbations that blend naturally and are harder to detect.}
    \label{fig:patches}
\end{figure*}

FourCastNet \cite{pathak2022fourcastnet} is evaluated on an ensemble of perturbed initial conditions. These initial conditions are generated by adding scaled Gaussian noise to the unperturbed state, where \(\delta_{ijk} \sim \mathcal{N}(0, 1)\). Specifically, different ensembles are created as \(\mathbf{Z}_0' = \mathbf{Z}_0 + \sigma \cdot \delta\) with \(\sigma = 0.3\). The resulting trajectories are produced as \(\phi(\mathbf{Z}_0')\). The paper shows that the ensemble mean predictions of FourCastNet closely follow the unperturbed control forecast, as assessed by the chosen evaluation metrics. This indicates that small Gaussian perturbations to the initial conditions do not cause significant deviations in the predictions, demonstrating the model's insensitivity to \emph{i.i.d.~random} minor variations in the initial weather fields.

However, to fully assess the robustness of FourCastNet to different inputs, it is essential to evaluate its performance under more targeted and potentially larger perturbations \(\delta\). We begin by constructing an attack aimed at solving optimization problem Equation (\ref{eqn:fundamental}), where no restrictions are placed on the initial perturbation:
\begin{equation}
\delta^{*}  = \min_{\delta \in \mathbb{R}^{L \times M \times N}} \left\| \phi_{T}(\mathbf{Z}_0 + \delta) - t_{\text{adv}} \right\|_2^2
\label{eqn:fundamental}
\end{equation}
The perturbation derived from solving Equation (\ref{eqn:fundamental}) influences all the prognostic variables listed in Table \ref{tab:prognostic_variables} (see Appendix), thereby altering the model’s initial conditions across multiple atmospheric fields. Figure \ref{fig:pointwise-mse} exemplifies this effect on the temperature field. In this test case, the model is tasked with predicting conditions 24 hours into the future, while the adversarial target $t_{\text{adv}}$ is set to reflect conditions at 120 hours. Such manipulations could yield serious real-world consequences. For instance, artificially elevated temperatures in the forecast might prompt authorities to prepare for a heatwave that will never materialize, potentially wasting resources and generating undue alarm. Conversely, underestimating temperatures could obscure an imminent heat risk, leaving communities unprepared and vulnerable to harm.

To visualize the distortion, Figure \ref{fig:pointwise-mse} compares the pointwise global temperature differences $\phi_{T}(\mathbf{Z}_0 + \delta^{*}) - t_{\text{adv}}$ and $\phi_{T}(\mathbf{Z}_0 + \delta^{*}) - \text{GT}$, where $\text{GT}$ is the true temperature field at 24 hours. Lighter regions correspond to smaller deviations, making it clear that the perturbed prediction adheres more closely to the adversarial target than to the authentic ground truth. To observe the individual temperature forecasts for the ground truth, unperturbed prediction, and perturbed prediction refer to  Figure \ref{fig:unconstrained_attack} (Appendix).

These results demonstrate that weather prediction models are inherently non-robust to arbitrary perturbations. However, such a \textit{naive} attack is impractical in real-world scenarios, as a meteorologist could readily detect discrepancies in the initial data by analyzing the divergent trajectories of various fields. In the following section, we investigate whether incorporating a layer of imperceptibility into the attack can achieve comparable results while enhancing the realism and practical applicability of these perturbations in controlled settings.

\section{Are Weather Forecast Models Secure to Localized Attacks on a Single Field?}

Building on the previous section, we extend our approach to ensure \textit{systemic stealthiness} in  \textbf{targeted} adversarial attacks inspired by the approach as in Equation \ref{eqn: target}. This is crucial because trajectory observers (e.g., meteorologist) may detect data corruption if the perturbed trajectories diverge significantly from expected behavior or if validation against ground truth fails during a pilot run. The objective is to identify a \textit{stealthy} perturbation, \(\delta\), that modifies the initial input \(\mathbf{Z}_0\) to create an adversarial example while adhering to stealth constraints. We impose stealthiness through the following criteria: First, the perturbation should affect \textbf{only a minimal subset of channels}, representing correlated variables, to align with the adversary's goal of targeting specific events (e.g., temperature) while limiting observable discrepancies. Second, the attack \textbf{must be localized to a specific spatial region}, such as simulating a heatwave warning confined to a targeted area. This adds to obfuscating the observable discrepancy for a chosen targeted variable. Third, the perturbation \textbf{must maintain physical realism} to avoid producing values that fall outside realistic bounds and could easily signal tampering. Finally,\textbf{ transitions introduced by the perturbation must be smooth}, avoiding abrupt or spiky changes that could highlight corrupted data from sensors. These constraints collectively ensure that adversarial perturbations remain both effective and undetectable within real-world operational settings.

To impose the channel sparsity constraint, only a subset of channels \(\mathbf{C}\), where \(\mathbf{C} \subseteq \{1, 2, \ldots, N\}\), is perturbed. For the localization constraint, the perturbations are confined to a spatial mask \(\mathbf{M}\).

To achieve these constraints, a projection operator \(\mathcal{P}_{\mathbf{M}, \mathbf{C}}\) is employed. The perturbed input \(\mathbf{Z}_0^\delta\) is defined as \(\mathbf{Z}_0 + \delta\), where \(\delta = \mathcal{P}_{\mathbf{M}, \mathbf{C}}(\delta)\). Each component of \(\delta\) is constrained such that channels not in \(\mathbf{C}\) are zeroed out, while those in \(\mathbf{C}\) are further localized by the mask \(\mathbf{M}\).

Additionally, the perturbations must generate forecast trajectories that are smooth and realistic, ensuring that the target fields remain within allowable limits. To impose these constraints, we utilize an adaptive loss objective that balances the primary objective with penalty terms for smoothness and realism. The primary objective, \(L_{\text{primary}}\), minimizes the difference between the perturbed forecast and the adversarial target \(t_{\text{adv}}\):
\[
L_{\text{primary}} = \left\| \phi_{T}(\mathbf{Z}_0 + \delta) - t_{\text{adv}} \right\|_2^2
\]

Two penalty terms encourage additional constraints: \(L_{\infty}\), which limits the maximum allowable values in the perturbed trajectories, and \(L_{\text{TV}}\), which minimizes the total variation to promote spatial smoothness. Each channel \(n \in \{1, \ldots, N\}\) is assigned a distinct penalty weight, \(\lambda_{\infty, n}\) and \(\lambda_{\text{TV}, n}\), along with maximum allowable limits, \(\epsilon_n\), and smoothness parameters, \(\tau_n\). As previously defined, \(\mathbf{Z}_t\) is the \(t^{\text{th}}\) prediction in a trajectory of length \(T\) which must obey these allowable and smoothness constraints:
\[
L_{\infty} = \sum_{t=0}^{T-1} \sum_{n=1}^{N} \lambda_{\infty, n} \cdot \max\left(0, \left\| (\mathbf{Z}_t)_n \right\|_\infty - \epsilon_n \right)
\]
\[
L_{\text{TV}} = \sum_{t=0}^{T-1} \sum_{n=1}^{N} \lambda_{\text{TV}, n} \cdot \max\left(0, \text{TV}\left( (\mathbf{Z}_t)_n \right) - \tau_n \right)
\]

The total loss function combines these terms to guide the optimization of \(\delta\), ensuring the perturbations are both realistic and stealthy:
\[
L(\delta) = L_{\text{primary}} + L_{\infty} + L_{\text{TV}}.
\]

To produce stealthy adversarial perturbations, we introduce \textbf{Weather Adaptive Adversarial Perturbation Optimization (WAAPO)}, summarized in Algorithm~\ref{alg:adversarial_perturbation}. WAAPO operates by iteratively updating the perturbation $\delta$ through gradient-based optimization. At each iteration, it computes the gradient of the total loss $L(\delta)$ with respect to $\delta$, which includes the primary objective $L_{\text{primary}}$ (ensuring the prediction aligns with the adversarial target $t_{\text{adv}}$) and two penalty terms, $L_{\infty}$ and $L_{\text{TV}}$, previously discussed, that collectively enforce physical realism, boundedness, and smoothness.

The algorithm then performs a gradient descent step to refine $\delta$. A subsequent projection step applies the channel and spatial masks $\mathbf{C}$  and $\mathbf{M}$, ensuring that only designated variables and regions are perturbed. By iterating this process until convergence or a fixed number of iterations is reached, WAAPO produces localized, channel-specific, and physically plausible adversarial examples.

\begin{algorithm}[H]
\footnotesize 
\caption{Weather Adaptive Adversarial Perturbation Optimization for Weather Forecast (WAAPO)}
\label{alg:adversarial_perturbation}
\begin{algorithmic}[1]
\Require Initial input $\mathbf{Z}_0$, target $t_{\text{adv}}$, learning rate $\alpha$, max iterations $K$, channels $\mathbf{C}$, spatial mask $\mathbf{M}$, penalties $\lambda_{\infty, n}$, $\lambda_{\text{TV}, n}$, constraints $\epsilon_n$, $\tau_n$
\Ensure Optimized perturbation $\delta_{WAAPO}$

\State Initialize perturbation: $\delta^{(0)} \gets \mathbf{0}$
\For{$k = 0$ to $K-1$}
    \State Compute Gradient: 
    \State $ \nabla_{\delta} L(\delta^{(k)}) \gets \frac{\partial}{\partial \delta} \left[ \| \phi_{T}(\mathbf{Z}_0 + \delta^{k}) - t_{\text{adv}} \|_2^2 \right. + $ 
    \State $ \left. \sum_{t=0}^{T-1} \sum_{n=1}^{N} \lambda_{\infty, n} \cdot \max\left(0, \| (\mathbf{Z}_t)_n \|_\infty - \epsilon_n \right) \right. + $ 
    \State $ \left. \sum_{t=0}^{T-1} \sum_{n=1}^{N} \lambda_{\text{TV}, n} \cdot \max\left(0, \text{TV}(\left( \mathbf{Z}_t)_n \right) - \tau_n \right) \right] $
    \State Gradient Descent Step: $\delta^{(k+1)} \gets \delta^{(k)} - \alpha \nabla_{\delta} L(\delta^{(k)})$
    \State Projection Step:
    \For{each channel $ n = 1 $ to $ N $}
        \If{$ n \in C $}
            \State $ \delta_n^{(k+1)} \gets \mathbf{M} \odot \delta_n^{(k+1)} $ \Comment{Apply spatial mask}
        \Else
            \State $ \delta_n^{(k+1)} \gets \mathbf{0} $ \Comment{Zero out perturbation}
        \EndIf
    \EndFor
\EndFor
\State \Return optimized perturbation $\delta_{WAAPO} =  \delta^{(K)}$
\end{algorithmic}
\end{algorithm}

\newpage
\section{Experimental Discussion}

\textbf{Hyperparameters}: 

We use samples from the ERA5 dataset for weather forecasting, starting from 2018, with data points spaced 6 hours apart. The initial condition \(\mathbf{Z}_{0}\) represents the first sample, and \(t_{adv}\) corresponds to the forecast 120 hours (5 days) after \(\mathbf{Z}_{0}\). Forecasting is performed using the FourCastNet model \cite{pathak2022fourcastnet}. The number of prognostic variables \(N\) is set to 20 (Table \ref{tab:prognostic_variables}). In experiments involving perturbation of specific channels, we perturb \(\mathbf{C} = \{\textbf{t2m}\}\), representing the temperature at 2 meters above the surface. The spatial mask for perturbations is configured with a patch starting at \((L_{0}, M_{0}) = (1100, 300)\) and a patch size of \((L_p, M_p) = (200, 300)\), chosen to target the South American region. The penalty parameters \(\lambda_{\infty, n}\) and \(\lambda_{\text{TV}, n}\) are set to 0.01, while the constraints \(\epsilon_n\) are determined as the maximum value of each field over a 5-day forecast. The smoothness parameter \(\tau_n\) is computed as the average smoothness value over the same period. Specific parameter values are detailed in the Appendix (Table \ref{tab:add_hyp}). Optimization is performed using the Adam optimizer with a learning rate of \(\alpha = 0.01\) and a total iteration count of \(K = 1000\).

\textbf{Performance of WAAPO}

\begin{figure*}[h]
        \centering
        
        \includegraphics[width=0.9\linewidth]{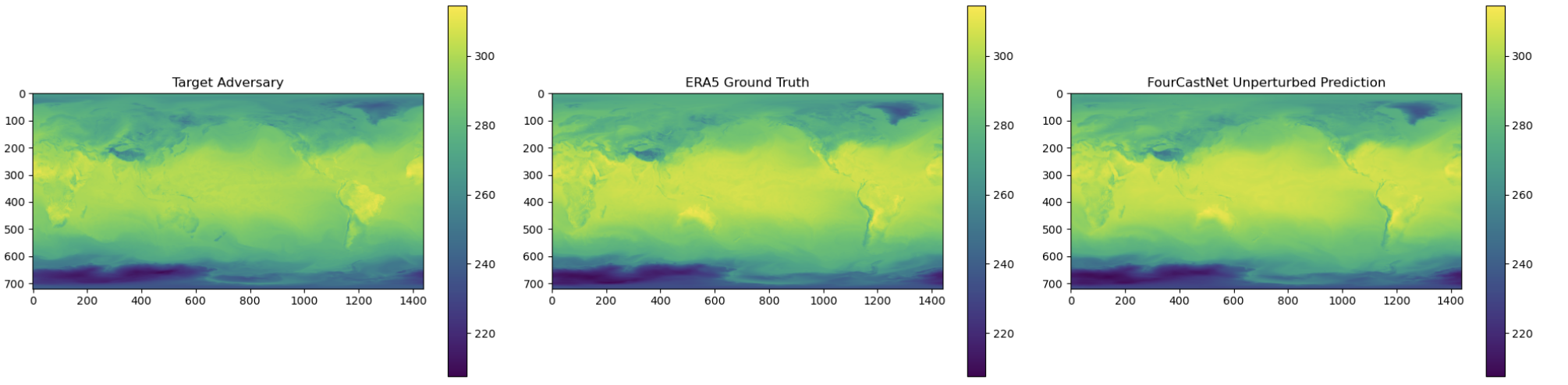}

        \includegraphics[width=0.9\linewidth]{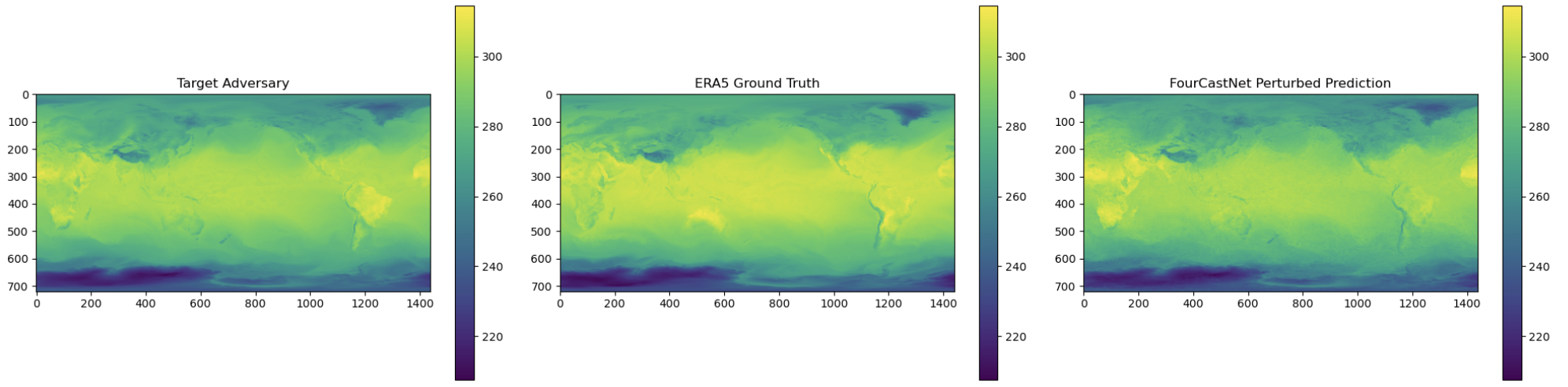}
        
        \vspace{0.3cm} 
        
        \caption{
            \textbf{Top:} Normal Prediction. 
            \textbf{Bottom:} Perturbed Prediction. 
            Illustration of the effect of the unconstrained adversarial attack on weather forecasts.
        }
        \label{fig:unconstrained_attack}
\end{figure*}

Utilizing the hyperparameters described, we solved for \(\delta^{*}\) in the unconstrained optimization setting in Equation \ref{eqn:fundamental}. However, solving for \(\delta_{WAPPO}\)required additional hyperparameter selection. Channel-based masking initially led to gradient explosions, resulting in a non-smooth, spiky loss trajectory that hindered optimizer convergence. To address this issue, we implemented norm-based gradient clipping to stabilize the gradients and incorporated a learning rate scheduler for dynamic adjustment. These techniques significantly enhanced optimizer stability, leading to improved convergence and more effective perturbation discovery. Consequently, we successfully induced temperature deviations without affecting other channels. Figure~\ref{fig:channel_only} illustrates the results when the field \textbf{t2m} (temperature at 2m above the surface, as described in Table~\ref{tab:prognostic_variables} in the Appendix) was only perturbed. It can be seen that the results are notably similar to \ref{fig:unconstrained_attack}.  Notably, even though some variables are correlated with \textbf{t2m}, e.g.~\textbf{U10}, \textbf{V10}, and \textbf{sp}, the perturbation was restricted solely to \textbf{t2m}, leaving other fields intact. \textbf{WAAPO} was thus able to successfully induce temperature deviations even while leaving correlated variables unchanged.

In a subsequent experiment, we applied localized temperature perturbations over South America to ensure that temperatures in other regions remained unaffected. This targeted perturbation involved applying a spatial mask to the perturbation, as shown in Figure~\ref{fig:patch_only}. The results indicate that the differences between the true forecast and the perturbed prediction are confined to the South American region, demonstrating the effectiveness of the spatial masking approach.

We also investigated the impact of the smoothness parameter on the perturbations. Figure~\ref{fig:patches} shows the results of localized perturbations with and without a smoothness penalty. As is evident from the figure, omitting the smoothness parameter produces coarse, visible patches, underscoring the importance of incorporating smoothness constraints. This comparison highlights the critical role of the smoothness parameter in ensuring realistic and physically consistent perturbations. Additionally, while the patch based attack is able to perturb the target region, the temperature profile of the affected region is still not similar to the target temperature profile. A possible reason could be the inherent spatial mixing that prevent effective execution of the targeted attack. In future work we explore how the adversary maybe able to utilize the potential inherent vulnerability of the process and improve \textbf{WAPPO} for the targeted localized patch based attacks.

While FourCastNet \cite{pathak2022fourcastnet} demonstrates robustness to random Gaussian noise added to its initial conditions, we observe that smaller, yet strategically crafted WAAPO perturbations can induce significant forecast errors. To quantify this effect, we introduce the Perturbation Magnitude Ratio relative to Gaussian (PMRG), which compares the Frobenius norm of $\delta_{\text{WAAPO}}$ to that of a scaled Gaussian perturbation, $\sigma \cdot \delta_{\text{Gaussian}}$, where $\delta_{\text{Gaussian}} \sim \mathcal{N}(0, 1)$ and $\sigma = 0.3$. The PMRG is defined as:

\[
\text{PMRG} = \frac{\|\delta_{\text{WAAPO}}\|_F}{\|\sigma \cdot \delta_{\text{Gaussian}}\|_F}
\]

indicates how the overall size of the WAAPO perturbation compares to a typical Gaussian disturbance of the same scale. Since 

\[
\|\sigma \cdot \delta_{\text{Gaussian}}\|_F \approx \sigma \sqrt{L \cdot M \cdot N},
\]

this provides a meaningful benchmark for evaluating magnitude differences.

\begin{table}[h!]
\centering
\begin{tabular}{@{}lcc@{}}
\toprule
\textbf{WAAPO Perturbation Variant} & \textbf{PMRG} \\ \midrule
$S_{\text{WAAPO}}^p$ (patch-based) & 0.105 \\
$S_{\text{WAAPO}}^c$ (channel-based) & 0.565 \\ 
\bottomrule
\end{tabular}
\caption{Comparison of WAAPO’s perturbation magnitudes relative to Gaussian noise. Despite being smaller, both variants substantially impact the model’s forecasts.}
\label{tab:ratios}
\end{table}

Table~\ref{tab:ratios} shows that both the patch-based ($S_{\text{WAAPO}}^p$) and channel-focused ($S_{\text{WAAPO}}^c$) attacks yield perturbations smaller than the baseline Gaussian noise. Yet, these ostensibly imperceptible adjustments—especially when localized to a single region or variable—can lead to significant deviations in FourCastNet’s predictions. This highlights a critical vulnerability: despite being smaller than random noise in magnitude, carefully targeted adversarial perturbations can exert a disproportionately large effect on AI-based weather forecasts.

\section{Conclusion }

In this work, we introduced \emph{Weather Adaptive Adversarial Perturbation Optimization } \textbf{(WAAPO)}, a novel framework for generating adversarial perturbations in weather forecasting models. By enforcing channel sparsity, spatial localization, and smoothness constraints, \textbf{WAAPO} produces perturbations that are imperceptible, physically valid, and localized to specific regions. Our experiments, conducted using the ERA5 dataset and FourCastNet \cite{pathak2022fourcastnet}, demonstrate that these carefully tailored attacks can align the model’s forecasts closely with adversarial targets—revealing critical vulnerabilities in forecasting systems that heavily depend on accurate initial conditions. As weather models increasingly inform decisions in agriculture, disaster management, and transportation, adversarial attacks that generate false weather alerts or hide extreme events pose significant risks, highlighting the need for effective safeguards.

Notably, our patch-based experiments show that while \textbf{WAAPO} can successfully perturb a targeted region, the resulting temperature profile does not fully match the adversarial target. A possible explanation is the inherent spatial mixing within the forecasting model, which dilutes localized perturbations over time. Future work will address this limitation by refining WAAPO’s methodology to exploit such mixing more effectively. In addition, the current study focuses mainly on the temperature field (\textbf{t2m}); expanding the attack space to other critical variables, such as surface pressure (\textbf{sp}), is essential for investigating impacts on hurricane forecasts and other severe weather scenarios. Beyond FourCastNet, evaluating adversarial robustness in alternative state-of-the-art models—including GraphCast \cite{lam2022graphcast}, ClimaX \cite{huang2023climax}, and PanguWeather \cite{bi2022pangu}—represents a crucial research direction. Finally, verification~\cite{xu2020automatic} will be vital in mitigating adversarial risks not only in weather forecasting, but also in broader climate modeling and environmental simulation tasks.

\section{Acknowledgements}
This work was performed under the auspices of the U.S. Department of Energy by Lawrence Livermore National Laboratory under Contract DE-AC52-07NA27344 and LDRD
Program Project No. 23-ERD-030 (LLNL-JRNL-871577).

\bibliography{aaai25}
\newpage
\section{Appendix}
\label{sec:Appendix}

\section{Real-World Adversarial Perturbations and Attack Strategies in Weather Forecasting Models}
\label{sec:perturbation}

In the context of weather forecasting models like FourCastNet, feasible adversarial attacks would aim to subtly alter the input variables to achieve a significant change in the forecast output. Consider some examples of perturbations and attack strategies that could be employed:

These are specific instances of the attack and we can keep the setting to be very generic.

\subsection*{1. Changing Hurricane Location}
\textbf{Objective:} Shift the predicted path of a hurricane to a different location.

\textbf{Perturbation:} Slightly adjust the wind speed, temperature, and pressure in the initial conditions around the hurricane’s current location to steer it towards a different path.

\subsection*{2. Altering Storm Intensity}
\textbf{Objective:} Modify the predicted intensity of a storm (e.g., making a hurricane appear stronger or weaker).

\textbf{Perturbation:} Small changes in sea surface temperature, pressure, and humidity around the storm's center to affect the storm's development and intensity.

\subsection*{3. Creating False Weather Events}
\textbf{Objective:} Generate forecasts that predict non-existent weather events, such as a hurricane or severe storm where none would occur.

\textbf{Perturbation:} Introduce small but spatially consistent perturbations across temperature, wind, and pressure fields in regions where no significant weather is expected to create the illusion of a developing storm.

\subsection*{4. Suppressing True Weather Events}
\textbf{Objective:} Prevent the model from predicting an actual severe weather event.

\textbf{Perturbation:} Apply small changes to key variables in the vicinity of the developing weather system to disrupt its formation in the model forecast.

\subsection*{5. (One Variable) Modifying Temperature Profiles}
\textbf{Objective:} Change the predicted temperature distribution across a region.

\textbf{Perturbation:} Adjust the initial temperature field to create warmer or cooler forecast conditions in targeted areas, which could affect predictions of heatwaves or cold spells.

\subsection*{6. (One Variable) Wind Pattern Manipulation}
\textbf{Objective:} Alter the predicted wind patterns, which could impact wind energy forecasts or general weather patterns.

\textbf{Perturbation:} Small, targeted changes to the initial wind field at various altitudes to influence the overall wind distribution.

\subsection*{7. Time Manipulation}

\textbf{Objective:} Change the predicted speed and timing of a hurricane's movement, causing delays or acceleration, and spread out the uncertainty in its path.

\textbf{Perturbation:} Adjust the initial conditions to alter the hurricane's speed and trajectory timing. This includes modifying wind speeds, pressure gradients, and other relevant atmospheric variables to delay or accelerate the hurricane's movement and increase the uncertainty in its predicted path

\subsection{Unconstrained Optimization}

\begin{table}[H]
\centering
\footnotesize 
\caption{Variables and Functions}
\begin{tabular}{|l|l|}
\hline
\textbf{Symbol} & \textbf{Description} \\ \hline
\(\mathbf{X}_0 \in \mathbb{R}^{N \times L \times M}\) & Original input \\ \hline
\(\delta \in \mathbb{R}^{N \times L \times M}\) & Perturbation \\ \hline
\(C \subseteq \{1, 2, \ldots, N\}\) & Channels to perturb \\ \hline
\(\mathbf{M} \in \mathbb{R}^{L \times M}\) & Spatial mask \\ \hline
\(\phi(\mathbf{Z}_t)_n\) & Model output at time \(t\) for channel \(n\) \\ \hline
\(\epsilon_n\) & Infinity norm constraint for channel \(n\) \\ \hline
\(\tau_n\) & Total variation constraint for channel \(n\) \\ \hline
\(\lambda_{\infty, n}, \lambda_{\text{TV}, n}\) & Penalty parameters for channel \(n\) \\ \hline
\(\alpha\) & Learning rate \\ \hline
\(\phi(\mathbf{Z}_{T-1} \mid \mathbf{X}_0^\delta)\) & Model output at final time step with perturbed input \\ \hline
\(\text{TV}(\cdot)\) & Total variation operator \\ \hline
\(\odot\) & Element-wise multiplication \\ \hline
\end{tabular}
\end{table}

\begin{table*}[h!]
    \centering
    \caption{20 Prognostic Variables Modeled by FourcastNet\cite{pathak2022fourcastnet}}
    \label{tab:prognostic_variables}
    \begin{tabular}{ll}
        \hline
        \textbf{Variable} & \textbf{Description} \\
        \hline
        \multicolumn{2}{l}{\textbf{Surface}} \\
        U10 & Zonal wind velocity at 10 meters above the surface \\
        V10 & Meridional wind velocity at 10 meters above the surface \\
        T2m & Temperature at 2 meters above the surface \\
        sp  & Surface pressure \\
        mslp & Mean sea level pressure \\
        \hline
        \multicolumn{2}{l}{\textbf{1000 hPa}} \\
        U1000 & Zonal wind velocity at 1000 hPa \\
        V1000 & Meridional wind velocity at 1000 hPa \\
        Z1000 & Geopotential height at 1000 hPa \\
        \hline
        \multicolumn{2}{l}{\textbf{850 hPa}} \\
        U850 & Zonal wind velocity at 850 hPa \\
        V850 & Meridional wind velocity at 850 hPa \\
        Z850 & Geopotential height at 850 hPa \\
        T850 & Temperature at 500 hPa \\
        R850 & Relative humidity at 850 hPa \\
        \hline
        \multicolumn{2}{l}{\textbf{500 hPa}} \\
        U500 & Zonal wind velocity at 500 hPa \\
        V500 & Meridional wind velocity at 500 hPa \\
        Z500 & Geopotential height at 500 hPa \\
        T500 & Temperature at 500 hPa \\
        R500 & Relative humidity at 500 hPa \\
        \hline
        \multicolumn{2}{l}{\textbf{50 hPa}} \\
        Z50 & Geopotential height at 50 hPa \\
        \hline
        \multicolumn{2}{l}{\textbf{Integrated Variables}} \\
        TCWV & Total Column Water Vapor \\
        \hline
    \end{tabular}
\end{table*}

\begin{figure}
    \centering
    \includegraphics[width=0.75\linewidth]{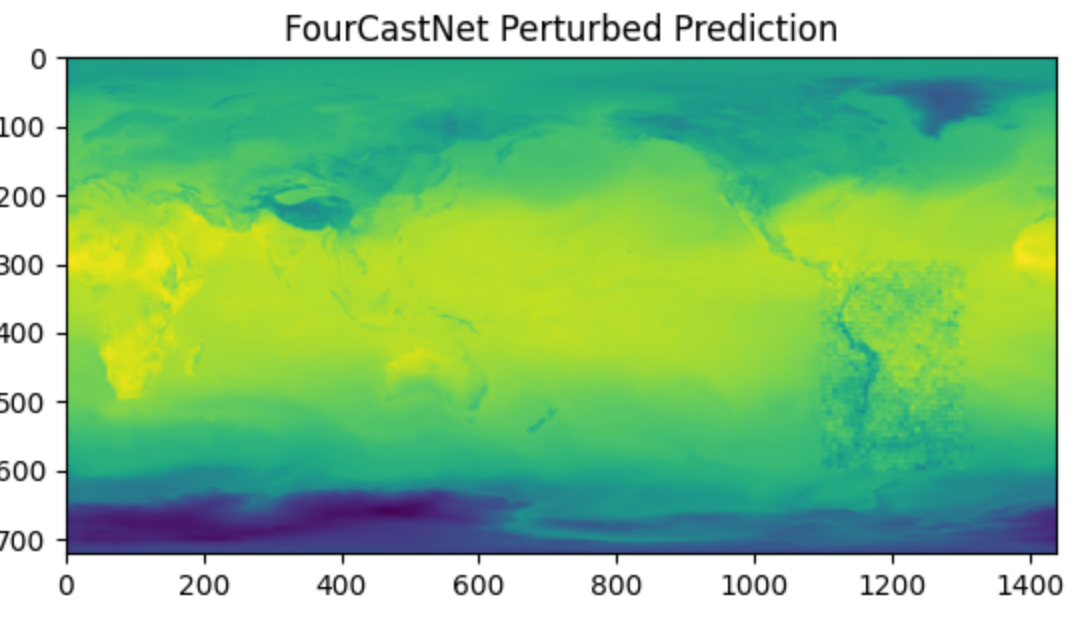}
    \caption{Coarse Patches.}
    \label{fig:coarse}
\end{figure}

\begin{table}[h!]
\centering
\begin{tabular}{|c|c|c|}
\hline
Channel \(n\) & $\epsilon_{n}$  & $\tau_{n}$  \\ \hline
1  & $9.1473 \times 10^{-8}$  & $99165.6016$   \\ \hline
2  & $2.5559 \times 10^{-6}$  & $109694.2422$  \\ \hline
3  & $1.3177 \times 10^{-6}$  & $27920.4980$   \\ \hline
4  & $2.8082 \times 10^{-7}$  & $58921.9922$   \\ \hline
5  & $1.8507 \times 10^{-6}$  & $34445.3281$   \\ \hline
6  & $4.2667 \times 10^{-7}$  & $22821.9902$   \\ \hline
7  & $1.2589 \times 10^{-7}$  & $98763.7188$   \\ \hline
8  & $5.4461 \times 10^{-7}$  & $108050.5078$  \\ \hline
9  & $1.9669 \times 10^{-6}$  & $32448.5762$   \\ \hline
10 & $5.8189 \times 10^{-7}$  & $89158.7656$   \\ \hline
11 & $1.7013 \times 10^{-6}$  & $102990.0625$  \\ \hline
12 & $3.5134 \times 10^{-6}$  & $20403.5078$   \\ \hline
13 & $3.3896 \times 10^{-7}$  & $61842.3750$   \\ \hline
14 & $1.8124 \times 10^{-7}$  & $63398.3281$   \\ \hline
15 & $2.4146 \times 10^{-6}$  & $13680.4805$   \\ \hline
16 & $1.8668 \times 10^{+0}$  & $19226.8984$   \\ \hline
17 & $3.6435 \times 10^{-6}$  & $9923.5332$    \\ \hline
18 & $4.8708 \times 10^{-6}$  & $114838.1641$  \\ \hline
19 & $1.6442 \times 10^{-6}$  & $142137.9844$  \\ \hline
20 & $6.4423 \times 10^{-8}$  & $49321.4219$   \\ \hline
\end{tabular}
\caption{Values of $\epsilon$  and $\tau$  for different channels.}
\label{tab:add_hyp}
\end{table}

\end{document}